# MODELS VS. INDUCTIVE INFERENCE FOR DEALING WITH PROBABILISTIC KNOWLEDGE

N. C. DALKEY*

## 1. Introduction

Two different approaches to dealing with probabilistic knowledge are examined--models and inductive inference. Examples of the first are: influence diagrams [1], Bayesian networks [2], log-linear models [3,4]. Examples of the second are: games-against nature [5,6], varieties of maximum-entropy methods [7,8,9], and the author's min-score induction [10].

In the modeling approach, the basic issue is manageability, with respect to data elicitation and computation. Thus, it is assumed that the pertinent set of users in some sense knows the relevant probabilities, and the problem is to format that knowledge in a way that is convenient to input and store and that allows computation of the answers to current questions in an expeditious fashion.

The basic issue for the inductive approach appears at first sight to be very different. In this approach it is presumed that the relevant probabilities are only partially known, and the problem is to extend that incomplete information in a reasonable way to answer current questions. Clearly, this approach requires that some form of induction be invoked. Of course, manageability is an important additional concern.

Despite their seeming differences, the two approaches have a fair amount in common, especially with respect to the structural framework they employ. Roughly speaking, this framework involves identifying clusters of variables which strongly interact, establishing marginal probability distributions on the clusters, and extending the subdistributions to a more complete distribution, usually via a product formalism. The product extension is justified on the modeling approach in terms of assumed conditional independence; in the inductive approach the product form arises from an inductive rule.

---

*Department of Computer Science, University of California, Los Angeles, Ca, 90024. This work was supported in part by National Science Foundation Grant IST 84-05161.

## 2. Structures on Event Spaces.

An event space is a set $X = X_1,\ldots,X_n$ of descriptors which is presumed to cover the subject matter of interest. For example, in a medical context, the $X$'s could be disease states, symptoms, test results, outcomes of treatment, and the like. Each descriptor $X_i$ involves a set of states $x_{ij}$, which is a partition (exclusive and exhaustive division) of the "universe" of potential cases. The vector $x=(x_1,\ldots,x_n)$ is a joint state for a specific case. It is presumed that there is a joint probability distribution $P(X)$ on the set of joint states, so that

$$\sum_X P(x) = 1$$
$$\sum_j P(x_{ij}) = 1$$

We define two types of components. An absolute component $Y$ is a subset of descriptors. A conditional component $(Z|W)$ is an ordered pair of absolute components. A probability $P(Y)$ on an absolute component is a joint distribution on the states $y$ for descriptors in $Y$; $P(Y)$ is a subdistribution (or marginal) of $P(X)$. Let $-Y$ denote the complement of $Y$ (all descriptors in $X$ not in $Y$). Thus

$$P(Y) = \sum_{-Y} P(X) \qquad (1)$$

A probability $P(Z|W)$ is a conditional probability distribution on the states $z$ given the states $w$. Thus

$$P(Z|W) = \frac{\sum_{-Z.W} P(X)}{\sum_{-W} P(X)} \qquad (1')$$

(The period in $-Z.W$ denotes the logical product "and".)

A set of components $C = Y_1,\ldots,Y_k$ is called a structure. The corresponding probability distributions on members of $C$ $PC = P(Y_1),\ldots,P(Y_k)$ is called a probability system (or system for short.) In this notation, $Y$ may be either absolute or conditional.

A system $PC$ is called consistent if there is a probability distribution $P(X)$ that fulfills (1) or (1') for all components $Y$ in $PC$. In general, if $PC$ is consistent, there will be a set $K(PC)$ of distributions compatible with $PC$.

In the model approach, it is assumed that a system $PC$ represents the clustering of descriptors with respect to dependence; i.e., within a component $Y$, the descriptors have

64

"strong" probabilistic interactions, whereas if $X_i$ and $X_j$ do not occur together in any component, then they are conditionally independent. Specifically,

$$P(X_i.X_j|-X_i.X_j) = P(X_i|-X_i.X_j)P(X_j|-X_i.X_j) \quad (2)$$

In the inductive approach, the system PC represents what is known concerning P(X). If $X_i$ and $X_j$ do not occur in a common component, then nothing is known about their probabilistic relationship.

A structure C by itself exhibits many of the general properties of the available knowledge. Thus, it is possible to determine that one structure is uniformly more informative than another, or to specify which structures have a product extension, without reference to the probabilities PC. [11] As the developers of influence diagrams and Bayesian networks have noted, this feature allows a significant amount of preliminary analysis to be carried out before numbers are introduced.

3. Webs

Structures have an internal organization; e.g., components may overlap. There are a number of different ways to represent this organization. A common representatation is as a graph, where the components are the nodes and one component is connected by an arc to another if they overlap. A more convenient representation for the purposes of this paper is what I call a web. Let [C] designate the set of all descriptors which belong to some member of C. An absolute component Y is terminal if Y consists of two subcomponents Z and W, Z.[C - {Y}] = 0, W ⊂ [C- {Y}]. Thus, W "connects" Z to the remainder of C. (W may be vacuous, in which case Z is unconnected to the remainder of C.) A conditional component Y = (Z|W) is terminal if the preceeding conditions hold for Z and W. (W cannot be vacuous for a conditional component.)

A web is a structure which fulfills the recursion:

1. Any absolute component is a web.

2. If Y is a terminal component, and C - {Y} is a web, then C is a web.

From the definition, a web contains at least one absolute component. A web can be "unpacked" to generate a linear order on the components by starting with any terminal component Y, labeling it 1, choosing any terminal component in C - {Y}, labeling it 2, and so on. A web is called conditional if all absolute components are distinct, i.e., do

65

not overlap.

A web is somewhat more general than influence diagrams or Bayesian networks. These can be characterized as conditional webs where for any conditional component $Y = (Z|W)$, $Z$ consists of a single descriptor.

Conditional webs are signicant for modeling probabilistic knowledge as a result of two basic properties:

a. The product $P^*(X) = \prod_C P(Y)$ is a joint probability distribution on $X$.

b. $P^*(X)$ is an extension of PC; i.e., it fulfills (1) or (1′) for all $Y$ in $C$.

Proofs for these assertions are readily constructed by induction on the number of components in a web.

What these two properties entail, in effect, is that if you can represent your knowledge concerning a distribution $P(X)$ by the sub-distributions PC for a web $C$ plus assuming conditional independence for descriptors not in common components, then the product $P(X)$ "automatically" expresses that knowledge.

From the modelling point of view, then, a web is a relatively manageable representation of probabilistic knowledge. All that need be input are the subdistributions PC. The product is quite convenient for computations; e.g., the manipulations feasible for influence diagrams are directly extendable to webs.

## 4. Induction and maximum-entropy

Turning to the inductive approach, in an earlier publication I demonstrated that for a subspecies of web, namely a forest, the product extension is the maximum entropy extension of PC. [11] A forest is a web in which all terminal components $Y = (Z:W)$ fulfill the additional restriction that $W$ is contained in some component $Y'$ in $C - \{Y\}$. (In a general web, $W$ need only be contained in the set of all descriptors "covered" by $C - \{Y\}$.) In the graphical representation mentioned earlier where arcs are defined by overlap of components, a forest is a graph with no loops. A forest corresponds to Goodman's decomposable model. [3]

Maximum entropy is an instance of min-score inference which has the dual properties: (a) guaranteed expectation--in the case of maximum entropy, the conclusion is always at least as informative as it claims to be--and (b) positive value of information--a conclusion based on additional

66

knowledge will be at least as informative as a conclusion without that knowledge. [10] Thus, if all you know is a set of subdistributions PC, and PC is a forest, then the product extension is a supportable estimate of the total distribution.

One of the motivations for studying webs was the expectation that the product extension would also turn out to be the maximum entropy extension for a general web. The expectation was based on a purported result of P. M. Lewis frequently cited in the literature to the effect that for a structure with a product extension, the product is the maximum entropy extension. [12] Unfortunately, the Lewis "result" happens to be incorrect.

An elementary counter-example is furnished by the simplest of all possible webs that is not a forest, namely the structure $C = \{X_1, X_2, (X_3|X_1.X_2)\}$. Set $P(X_1) = P(X_2) = .5$ and define $P(X_3|X_1.X_2)$ by Table I, where "1" means occurrance and "0" means non-occurrance in the list of cases.

Table I

| $X_1$ | $X_2$ | $P(X_3|X_1.X_2)$ |
|---|---|---|
| 1 | 1 | 1 |
| 1 | 0 | .5 |
| 0 | 1 | .5 |
| 0 | 0 | 0 |

The product distribution $P^\bullet(X)$ is displayed in Table II, along with another distribution $P^o(X)$. $P^o(X)$ is an extension of PC--which can easily be verified by summation--and is also clearly a higher entropy distribution.

Table II

| $X_1$ | $X_2$ | $X_3$ | $P^\bullet(X)$ | $P^o(X)$ |
|---|---|---|---|---|
| 1 | 1 | 1 | .25 | 1/6 |
| 1 | 1 | 0 | 0 | 0 |
| 1 | 0 | 1 | .125 | 1/6 |
| 1 | 0 | 0 | .125 | 1/6 |
| 0 | 1 | 1 | .125 | 1/6 |
| 0 | 1 | 0 | .125 | 1/6 |
| 0 | 0 | 1 | 0 | 0 |
| 0 | 0 | 0 | .25 | 1/6 |

The entropy of $P^\bullet(X) = - \sum_X P^\bullet(x) \log P^\bullet(x) = 1.7329$, whereas the entropy of $P^o(X) = 1.7918$. The numerical differ-ence in entropy is small, but the difference between .25 and 1/6 for P(1,1,1), e.g., may not seem trivial.

The elementary structure C of the example is actually a substructure of any web that is not a forest. Hence a similar counter-example can be constructed for any such web. The example is also a counter to the Lewis "result'.

The upshot of this inquiry, then, is that a forest is the most general structure for which the product extension is always the maximum entropy extension.

## 5. Discussion

At first glance, the fact that the product extension of a web is not in general maximum entropy may appear benign. From the standpoint of the model approach, the basic properties of a web--the product is a probability and an extension of PC--make webs a highly convenient representation of probabilistic knowledge. All that is lost is a desirable, but by no means essential, fallback. In the case of a forest, for example, if the assumption of conditional independence for separated descriptors is shaky, then it can still be contended that the product is a reasonable estimate of the joint distribution, given PC. It would be a valuable safety feature if the same could be claimed for a web.

From the standpoint of the inductive approach, it is perhaps unfortunate that the product extension of a web is not maximum entropy. However, the maximum entropy extension can be sought by other means. [13] What is lost is the convenience of the product form. For the complex systems of many descriptors common in expert systems, maximum entropy formalisms are likely to be cumbersome.

On a somewhat deeper level, however, the result is thought-provoking. Independence is a common "simplifying" assumption in expert systems. [14] The maximum entropy property, where germaine, is a good justification of the "assumption" even when there is no evidence either for or against independence. However, as the example shows, maximum entropy does not imply independence, not even conditional independence, if the structure is not a forest. In the example, $P^{\circ}(x_1|x_3) = P^{\circ}(x_2|x_3) = 2/3$; but $P^{\circ}(x_1.x_2|x_3) = 1/3$, rather than 4/9 as required by conditional independence.

One route that can be taken is to "prune" the structure to a forest. Lemmer [15] has adopted this suggestion, following a program proposed by Lewis [12], Chow and Liu [16], and others. The advantages of this approach are clear: substantive inputs can be restricted to the subdistributions in PC for the forest, the product extension is automatically consistent with the inputs, and, as I mentioned above, the

68

fact that the product extension is maximum entropy carries
strong guarantees.

A basic element missing from this program is a measure
of the information that is lost by the pruning process.
Concommitantly, there is no systematic procedure for deter-
mining the most informative forest contained in the knowledge
available to the analyst. Given a general probability system
PC, if PC is consistent, the amount of information in PC can
be defined as $\max_{P \in K(PC)} \text{Entropy}(P)$. At present, there is no
way to determine this quantity directly from PC--or, for that
matter, determining whether PC is consistent. These issues
appear to be one area of potentially fruitful research.